%% file: a_final.tex
\documentclass[10pt,twocolumn,letterpaper]{article}
\pdfoutput=1
\usepackage{iccv}
\usepackage{times}
\usepackage{epsfig}
\usepackage{graphicx}
\usepackage{amsmath}
\usepackage{amssymb}
\usepackage{multirow}
\usepackage{siunitx}
\usepackage{soul}
\usepackage{subfiles}
\usepackage{graphicx}
\usepackage{subcaption}
\usepackage[accsupp]{axessibility} 

\newcommand{\argmax}{\text{argmax}}
\newcommand{\softmax}{\text{softmax}}
\newcommand{\sg}{\text{sg}}
\newcommand{\nonzero}{\text{nonzero}}

\usepackage[breaklinks=true,bookmarks=false]{hyperref}

\hypersetup{
    colorlinks=true, 
    urlcolor=blue,
}

\iccvfinalcopy 

\def\iccvPaperID{12726} 

\ificcvfinal\fi

\begin{document}

\title{Improving Representation Learning for Histopathologic Images with Cluster Constraints}

\author{
   Weiyi Wu$^{1}$  \ \ \ \ 
   Chongyang Gao$^{2}$  \ \ \ \ 
   Joseph DiPalma$^{1}$ \ \ \ \ 
   Soroush Vosoughi$^{1}$ \ \ \ \ 
   Saeed Hassanpour$^{1}$ \\\\
   $^1$Dartmouth College  \ \ \ \ \  $^{2}$Northwestern University\\\\
    \textnormal{{\{weiyi.wu.gr,joseph.r.dipalma.gr,soroush.vosoughi,saeed.hassanpour\}@dartmouth.edu}}\\
    \textnormal{chongyanggao2026@u.northwestern.edu}
}

\maketitle
\ificcvfinal\thispagestyle{empty}\fi
\begin{abstract}

Recent advances in whole-slide image (WSI) scanners and computational capabilities have significantly propelled the application of artificial intelligence in histopathology slide analysis. While these strides are promising, current supervised learning approaches for WSI analysis come with the challenge of exhaustively labeling high-resolution slides—a process that is both labor-intensive and time-consuming. In contrast, self-supervised learning (SSL) pretraining strategies are emerging as a viable alternative, given that they don't rely on explicit data annotations. These SSL strategies are quickly bridging the performance disparity with their supervised counterparts. In this context, we introduce an SSL framework. This framework aims for transferable representation learning and semantically meaningful clustering by synergizing invariance loss and clustering loss in WSI analysis. Notably, our approach outperforms common SSL methods in downstream classification and clustering tasks, as evidenced by tests on the Camelyon16 and a pancreatic cancer dataset. The code and additional details are accessible at: \url{https://github.com/wwyi1828/CluSiam}.

\end{abstract}

\subfile{./introduction.tex}

\subfile{./related_works.tex}
\subfile{./method.tex}
\subfile{./experiments_results.tex}

\section{Conclusion}

In this paper, we introduce CluSiam, a SSL technique that integrates cluster constraints to enhance representation learning for histopathology images. By subtly pushing apart inter-cluster instances while aligning intra-cluster views, CluSiam balances similarity and dissimilarity. It demonstrates substantial improvements in downstream classification and clustering tasks compared to baseline methods. Additionally, CluSiam provides an efficient way to analyze histopathology images without requiring manual annotations.

\section{Acknowledgements}
This research was supported in part by grants from the US National Library of Medicine (R01LM012837 and R01LM013833) and the US National Cancer Institute (R01CA249758).

\clearpage

{\small
\bibliographystyle{ieee_fullname}
\bibliography{egbib}
}

\end{document}


\title{Supplementary Material for \\ Improving Representation Learning for Histopathologic Images with Cluster Constraints}

\author{
   Weiyi Wu$^{1}$  \ \ \ \ Chongyang Gao$^{2}$  \ \ \ \ Joseph DiPalma$^{1}$ \ \ \ \ Soroush Vosoughi$^{1}$ \ \ \ \ Saeed Hassanpour$^{1}$ \\
   $^1$Dartmouth College  \ \ \ \ \  $^{2}$Northwestern University
}

\maketitle



\section{Implementation Details}

This section provides details on the implementation of various self-supervised learning methods. The hyper-parameters and data augmentations for these methods are summarized below.

\subsection{SimCLR}
We used settings similar to those in the commonly used SimCLR repository~(\url{https://github.com/sthalles/SimCLR}).

We configured the SimCLR model using the following hyper-parameters and augmentations:

\begin{itemize}
\item Optimizer: Adam
\item Scheduler: Cosine Annealing
\item Learning Rate: 0.1
\item Weight Decay: 1e-4
\item Batch Size: 512
\item Crop: Scale = (0.08, 1.0)
\item ColorJitter: Scale = (0.8,0.8,0.8,0.2), Probability = 0.8
\item Grayscale: Probability = 0.2
\item GaussianBlur: Scale = (0.1,2), Probability = 0.5
\item Horizontal Flip: Probability = 0.5
\end{itemize}

\subsection{SwAV}
We used settings similar to those in the official SwAV repository~(\url{https://github.com/facebookresearch/swav/}). 

\begin{itemize}
\item Optimizer: SGD
\item Scheduler: Cosine Annealing
\item Learning Rate: 0.1
\item Weight Decay: 1e-4
\item Batch Size: 512
\item Momentum: 0.9
\item Crop: Scale = (0.2, 1.0)
\item ColorJitter: Scale = (0.4,0.4,0.4,0.1), Probability = 0.8
\item Grayscale: Probability = 0.2
\item GaussianBlur: Scale = (0.1,2), Probability = 0.5
\item Horizontal Flip: Probability = 0.5
\end{itemize}

\subsection{PCL}
We used settings similar to those in the official PCL repository~(\url{https://github.com/salesforce/PCL/})


\begin{itemize}
\item Optimizer: SGD
\item Scheduler: Cosine Annealing
\item Learning Rate: 0.03
\item Weight Decay: 1e-4
\item Batch Size: 512
\item Momentum: 0.9
\item Crop: Scale = (0.2, 1.0)
\item ColorJitter: Scale = (0.4,0.4,0.4,0.1), Probability = 0.8
\item Grayscale: Probability = 0.2
\item GaussianBlur: Scale = (0.1,2), Probability = 0.5
\item Horizontal Flip: Probability = 0.5
\end{itemize}

\subsection{Barlow Twins}
We used settings similar to those in the official Barlow Twins repository~(\url{https://github.com/facebookresearch/barlowtwins}). We present two different augmentation settings here, as Barlow Twins uses asymmetric augmentation techniques for its two views.

\begin{itemize}
\item Optimizer: LARS
\item Scheduler: Cosine Annealing
\item Base Learning Rate: 2
\item Weight Decay: 1e-6
\item Batch Size: 512
\end{itemize}
Augmentation 1:
\begin{itemize}
\item Crop: Scale = (0.08, 1.0)
\item ColorJitter: Scale = (0.4,0.4,0.2,0.1), Probability = 0.8
\item Grayscale: Probability = 0.2
\item GaussianBlur: Scale = (0.1, 2.0), Probability = 1.0
\item Solarization: Probability = 0.0
\item Horizontal Flip: Probability = 0.5
\end{itemize}
Augmentation 2:
\begin{itemize}
\item Crop: Scale = (0.08, 1.0)
\item ColorJitter: Scale = (0.4,0.4,0.2,0.1), Probability = 0.8
\item Grayscale: Probability = 0.2
\item GaussianBlur: Scale = (0.1, 2.0), Probability = 0.1
\item Solarization: Probability = 0.2
\item Horizontal Flip: Probability = 0.5
\end{itemize}

\subsection{BYOL}
We used settings similar to those in the widely-used BYOL repository~(\url{https://github.com/lucidrains/byol-pytorch}).
\vspace{1cm}
\begin{itemize}
\item Optimizer: Adam
\item Learning Rate: 3e-4
\item Batch Size: 512
\item Crop: Scale = (0.08, 1.0)
\item ColorJitter: Scale = (0.8,0.8,0.8,0.2), Probability = 0.8
\item Grayscale: Probability = 0.2
\item GaussianBlur: Scale = (1,2), Probability = 0.2
\item Horizontal Flip: Probability = 0.5
\end{itemize}

\subsection{SimSiam \& CluSiam}
In the SimSiam and corresponding CluSiam training, we used settings similar to those in the official SimSiam repository~(\url{https://github.com/facebookresearch/simsiam}).

\begin{itemize}
\item Optimizer: SGD
\item Scheduler: Cosine Annealing
\item Learning Rate: 0.1
\item Weight Decay: 1e-4
\item Batch Size: 512
\item Momentum: 0.9
\item Fix Prediction Learning Rate: True
\item RandomCrop: Scale = (0.2, 1.0)
\item ColorJitter: Scale = (0.4,0.4,0.4,0.1), Probability = 0.8
\item Grayscale: Probability = 0.2
\item GaussianBlur: Scale = (0.1,2), Probability = 0.5
\item Horizontal Flip: Probability = 0.5
\end{itemize}

\subsection{Supervised Learning}
We followed settings similar to those in SimSiam/CluSiam to train our supervised model. A weighted loss function was used to account for class imbalance in the patch dataset. 


\begin{itemize}
\item Optimizer: Adam
\item Scheduler: Cosine Annealing
\item Learning Rate: 0.1
\item Weight Decay: 1e-4
\item Batch Size: 512
\item Weighted Loss: True
\item Crop: Scale = (0.2, 1.0)
\item ColorJitter: Scale = (0.4,0.4,0.4,0.1), Probability = 0.8
\item Grayscale: Probability = 0.2
\item GaussianBlur: Scale = (0.1,2), Probability = 0.5
\item Horizontal Flip: Probability = 0.5
\end{itemize}

%% file: introduction.tex
\section{Introduction}

Histopathology slide analysis remains the gold standard for cancer diagnosis and prognosis. In recent years, researchers have seen the burgeoning adoption of digital histopathology slides in pathology laboratories, thanks to the availability of digital pathology scanners and advancements in computer vision, revolutionizing computational pathology~\cite{janowczyk2016deep}. While adoption of digital slides has accelerated, progress has been hindered by the exceptionally high resolution of whole slide images (WSIs), often exceeding $40,000\times40,000$ pixels, which makes directly applying standard computer vision models to WSIs not feasible. Furthermore, downsampling WSIs to a more manageable magnification level results in a substantial loss of fine-grained visual information.

\begin{figure}[!ht]
\center
\includegraphics[width=\linewidth]{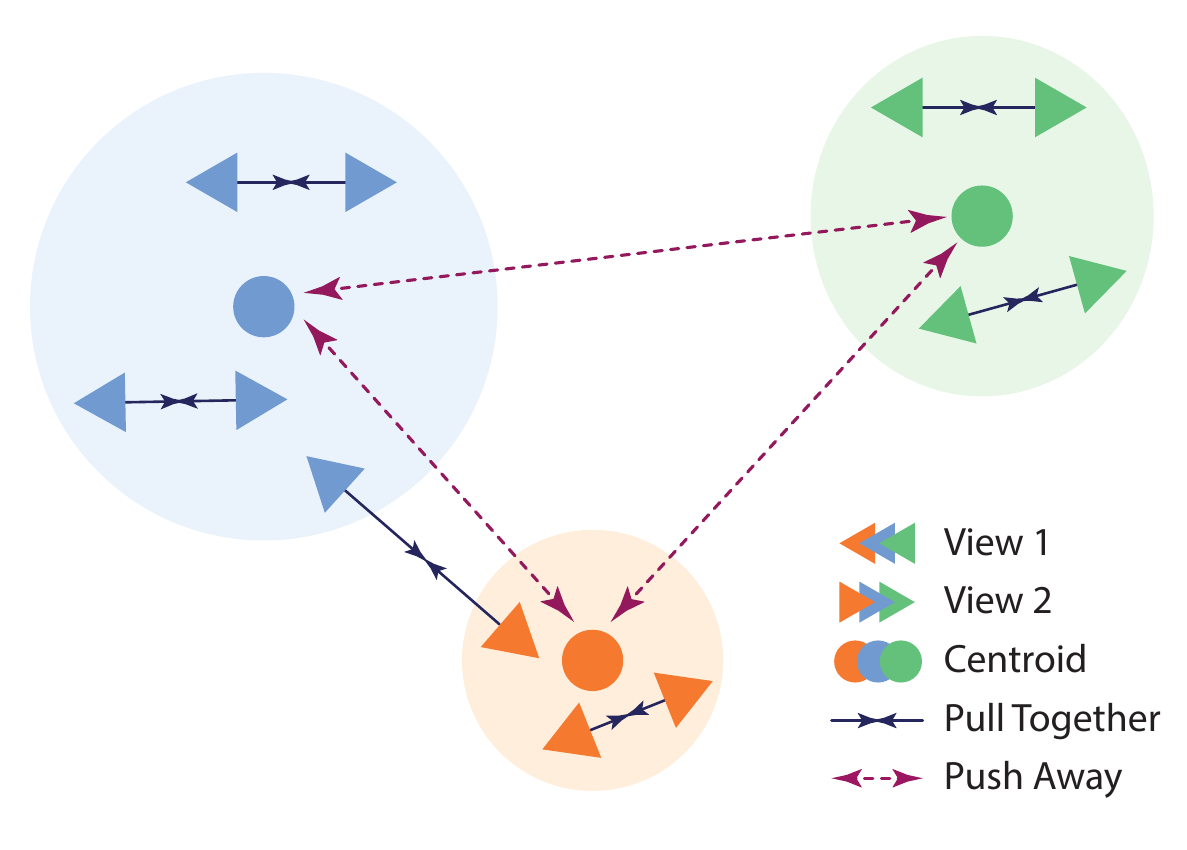}
\caption{The framework of CluSiam. View 1 and View 2 are distinct augmentations of the same images, pooled together for clustering. The invariance loss (solid line) aligns representations of the two views, while the cluster loss (dashed line) pushes cluster centroids apart.}
\label{fig:overview}
\end{figure}

To address the challenges, WSIs are commonly subdivided into more manageable patches through sliding-window techniques. These patches are then labeled using annotations, forming training data for a patch-level classifier. Features extracted by the trained patch-level classifiers are aggregated to infer slide-level label~\cite{hou2016patch,wei2019pathologist,wu2023graph,wang2016deep,liu2017detecting}. However, this annotation-reliant approach has a significant drawback. It's heavily dependent on precise annotations, which are expensive to obtain. Annotating WSIs is a painstaking and error-prone task that requires pixel-by-pixel scrutiny from highly skilled pathologists. The elusive borders between different tissue patterns introduce variability among pathologists. Additionally, tissue morphology's inherent variability often further diminishes the accuracy of annotations. Therefore, obtaining precise and consistent annotations remains an uphill battle, even with substantial expertise and time invested by trained pathologists. Inaccurate annotations can potentially lead to inaccurate and inconsistent WSI analysis models~\cite{wahab2022semantic,ashraf2022loss}.

To mitigate the impact of inaccurate annotations, noise-aware learning models have emerged. These methods improve the performance of patch-level feature extractors by either filtering or down-weighting the noisy patches~\cite{ashraf2022loss,le2019pancreatic,cheng2020self}. However, these models are still constrained by the annotation bottleneck. Even acquiring noisy annotations for WSIs demands substantial time and expertise, which motivates the need for annotation-free techniques. They reduce costs and save time. They also eliminate the impact of inaccurate annotations.

In this challenging landscape, annotation-free techniques have emerged as a promising solution. By requiring only whole-slide labels, they not only cut costs and save time but also eliminate the effects of annotation inaccuracies. Among these, Chen \etal \cite{chen2021annotation} proposed a method that leverages a unified memory mechanism to train convolutional neural networks (CNNs) directly with numerous images. However, this approach is constrained to lower magnification levels, restricting pixel sizes to above \SI{2}{\micro\meter}. Conversely, other studies~\cite{liu2017detecting, li_dual-stream_2021, d2022comparison, dipalma2021resolution} have shown that achieving better results is possible by employing higher or multi-scale magnification levels across a variety of model designs.

Weakly supervised techniques have gained popularity as an annotation-free approach that retains high-resolution details by utilizing slide-level labels instead of exhaustive patch-level annotations. Obtaining slide-level labels is less laborious compared to exhaustive patch-level annotations. Thus, weakly supervised learning has become particularly popular for histology slide classification tasks~\cite{10.1001/jamanetworkopen.2019.14645,tomita2022predicting,zhao2020predicting}. These methods employ slide-level labels as weak supervision for all patches within a slide. Multiple instance learning (MIL) models leverage this by treating slides as positive or negative bags, with patches as instances~\cite{lerousseau2020weakly,li_dual-stream_2021,ilse2018attention,zhao2020predicting,campanella2019clinical,mercan2017multi}. However, MIL models have some limitations. They often neglect important contextual cues across a whole slide. Additionally, off-the-shelf feature extractors pretrained on natural images fail to sufficiently capture tissue morphology. These drawbacks motivate exploring self-supervised approaches for histology slides.

Self-supervised learning (SSL) enables models to learn feature representations without the need for labels. SSL methods are rapidly closing the performance gap with supervised approaches. However, SSL typically requires a large sample size. but this is mitigated for high-resolution histopathology images by splitting WSIs into numerous small patches. In computational pathology, self-supervision methods become an appealing solution for annotation-free WSI analysis \cite{li_dual-stream_2021, kang2023benchmarking,jiang2023masked, yang2022concl, carse2021unsupervised}. These methods utilize multiple-instance learning to aggregate self-supervised patch representations. They have demonstrated the capability to match the performance of state-of-the-art supervised methods while reducing the annotation burden on pathologists by eliminating the need for manual annotations.

One of the key paradigms of SSL is contrastive-based SSL~\cite{he2020momentum, chen2020simple,oord2018representation, van2020scan, jian2022non}. They may not be the most effective in histopathology image analysis because adjacent patches from a WSI can be very similar in their morphological features, making them unsuitable as negative sample pairs. These methods also rely on a large number of negative pairs. To avoid the need for negative pairs, some knowledge-distillation-based methods \cite{chen2021exploring,caron2021emerging,grill2020bootstrap} concentrate solely on positive sample pairs, which are defined using augmented views of the same image. However, only focusing on positive pairs might prevent them from learning global information, as their objective functions only consider augmentations from the same image.

Apart from SSL representation learning, another pivotal technique gaining attention is clustering. Clustering is an unsupervised learning approach where similar samples are grouped to ensure intra-cluster cohesion and inter-cluster separation. In the domain of WSI retrieval, clustering could be instrumental. Wang \etal \cite{wang2023retccl} employed a K-Means clustering-driven architecture, while Chen \etal \cite{chen2022fast} integrated a self-supervised variational autoencoder with the K-Means algorithm, both for WSI retrieval systems. Given the growing prominence of such methods in WSI retrieval, there's an increasing demand to refine these clustering algorithms within computational pathology. Clustering shares similarities with representation learning. This has inspired clustering-based SSL methods that use pseudo-labels from iterative K-Means clustering algorithms for training feature encoders. Although these methods can learn effective image representations, they may not improve the performance of the actual clustering tasks as they cluster images into thousands of groups, which might hinder their direct use for histopathology image retrieval. Large cluster counts make identifying relevant groups challenging.

To address the shortcomings, we propose Cluster-Siam (CluSiam), a framework that decouples clustering from representation learning and retains only the most relevant and interpretable clusters for medical applications. CluSiam takes advantage of an existing self-supervised backbone to extract representations. We introduce a cluster loss to guide the backbone in learning effective representations while generating accurate, interpretable cluster assignments for histopathology images. (Figure~\ref{fig:overview}). Our experiments demonstrate that CluSiam outperforms baselines on downstream classification tasks. Additionally, our adaptive clustering algorithm outperforms K-Means in clustering, resulting in improved cluster assignments. In addition, our cluster assigner emerges as a by-product of the representation learning process, thus introducing only a small additional computational cost once the training is complete.

\begin{figure}[!ht]
    \center
    \includegraphics[width=\linewidth]{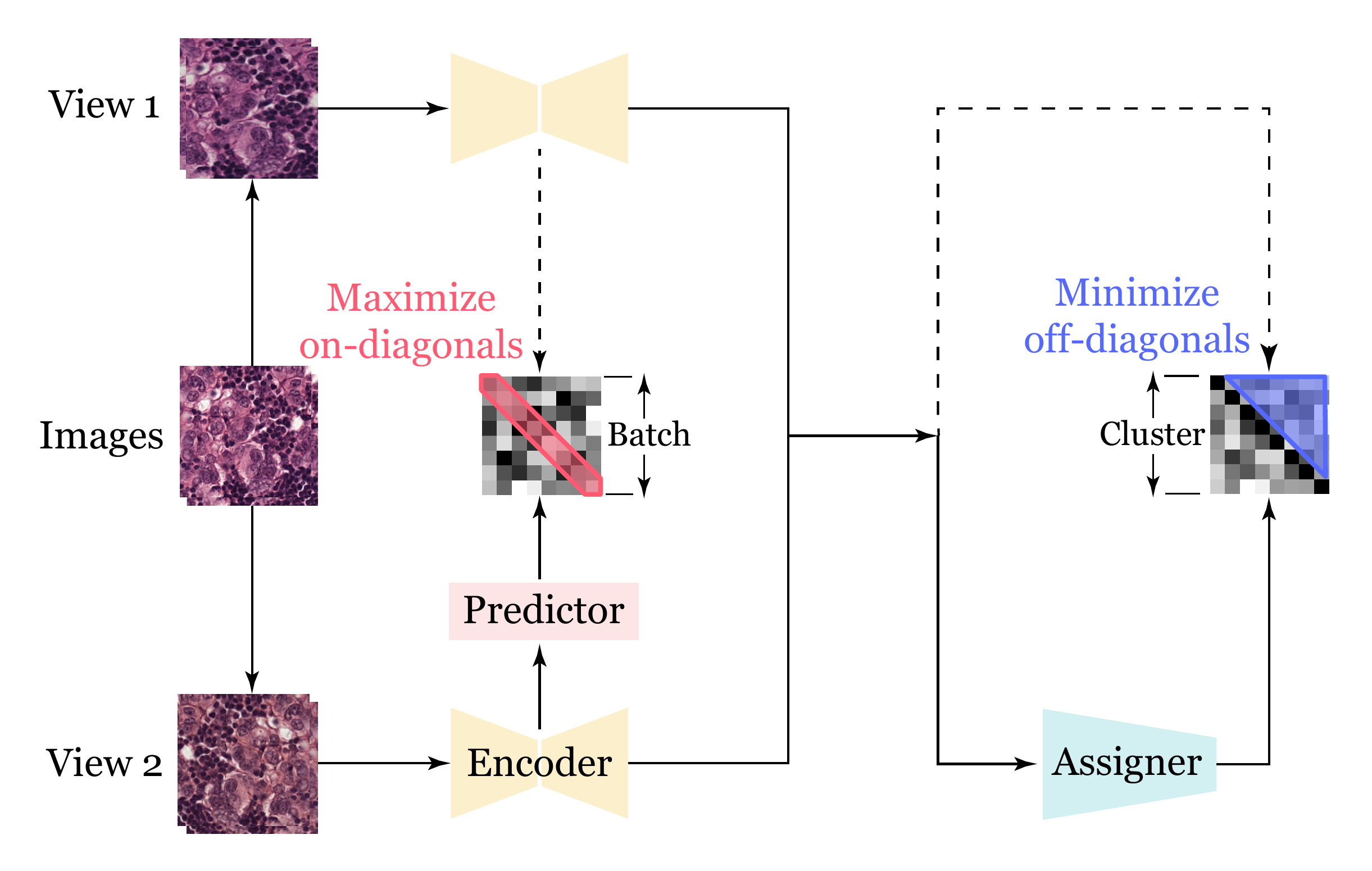}
    \caption{\label{schema}The details of the CluSiam framework. The invariance loss maximizes the on-diagonal elements of the similarity matrix between views. The cluster assigner takes the concatenated views as input and generates cluster assignments. The cluster loss minimizes the off-diagonal elements of the similarity matrix between cluster centroids.}
    \label{fig:detail}
  \end{figure}

The contributions of this paper can be summarized as follows:
\begin{itemize}
\item[•] We propose CluSiam, a SSL framework for image representation learning and clustering that combines invariance loss and cluster loss (Figure~\ref{fig:detail})
\item[•] We compare the performance of different SSL frameworks and demonstrate that CluSiam outperforms popular SSL methods on multiple histopathology datasets.
\item[•] CluSiam provides an efficient and accurate way to cluster histopathology images without either patch-level annotations or slide-level labels, with clustering performance better than the widely used K-Means clustering in digital pathology.
\end{itemize}

%% file: related_works.tex
\section{Related Works}

\subsection{Self-supervised learning}

Self-supervised learning (SSL) methods have recently demonstrated effectiveness for computer vision tasks by learning representations without reliance on manual labels. SSL techniques leverage the intrinsic structure and consistency of the data itself as a supervisory signal. Several paradigms have arisen, including contrastive-based, knowledge-distillation-based, clustering-based, and information maximization-based approaches. Typically, these techniques function by generating augmented pairs of views from a single data instance and directing the model to produce similar outputs for each view.

Contrastive learning stands out as a key self-supervised approach in representation learning, with the goal of deriving informative and concise representations from unstructured data. A slew of methods grounded in contrastive learning have been proposed, including Contrastive Predictive Coding (CPC) \cite{oord2018representation}, SimCLR \cite{chen2020simple}, MoCo \cite{he2020momentum}, and NNCLR \cite{dwibedi2021little}. CPC, recognized for its widespread application, employs an autoregressive model to predict future observations based on past observations, rendering it particularly apt for sequential data. MoCo and SimCLR, two other popular contrastive learning techniques for instance discrimination, generate positive pairs by utilizing two different views (augmentations) of the same image and negative pairs by pairing augmentations of different data points. The principle behind contrastive learning is to distance negative pairs while converging positive pairs. However, achieving optimal performance with these methods often necessitates a plethora of negative pairs. MoCo addresses this issue by implementing momentum encoders and a memory bank mechanism, while SimCLR capitalizes on large batch sizes for negative pair comparisons. In NNCLR, the model learns representations by minimizing the distance between an anchor and its nearest neighbor in the momentum encoder's output space while maximizing the distance to other negative samples. This method streamlines the utilization of negative samples in a batch, curtailing the need for large batch sizes and memory banks while maintaining competitive performance compared to other contrastive learning methods.

Knowledge-distillation-based methodologies, such as BYOL \cite{grill2020bootstrap} and SimSiam \cite{chen2021exploring}, aim to enhance performance with smaller batch sizes and without the need for negative samples. In stark contrast to their contrastive counterparts, these non-contrastive techniques employ only positive pairs, eliminating the demand for large batch sizes or a memory bank mechanism. BYOL stands out with its momentum update mechanism, which renders negative pairs unnecessary. It establishes a target network by applying an exponential moving average to the online network's weights. This "moving target" offers the online network a stable benchmark to aim for during training, pushing the network away from trivial solutions and encouraging richer representations. SimSiam, influenced by BYOL, streamlines the process by forgoing the moving target. Instead, it adopts a symmetric architecture, where dual networks reciprocally predict each other's outputs. To ensure the representations are non-trivial, SimSiam utilizes a stop-gradient operation. Nonetheless, despite their batch efficiency, both BYOL and SimSiam remain susceptible to collapsing into trivial solutions.

Clustering-based SSL methods, such as DeepCluster \cite{caron2018deep} and Prototypical Contrastive Learning (PCL) \cite{liprototypical}, offer an innovative angle to representation learning by capitalizing on iterative pseudo-labeling to cluster the learned representations. DeepCluster updates its network parameters according to pseudo-labels produced by the K-Means clustering algorithm. This aligns the network more closely with the inherent data distribution. Expanding on the foundation laid by MoCo, PCL integrates the ProtoNCE loss and K-Means clustering, aiming to refine image embeddings by nudging them closer to their respective prototypes by optimizing the ProtoNCE loss function. SwAV \cite{caron2020unsupervised}, another clustering-based SSL method, shares similarities with SimSiam architecture but differentiates itself with a swap prediction mechanism. Specifically, SwAV aligns the cluster assignments of one augmentation with the representations of another version of the same image. These assignments are fine-tuned using the Sinkhorn algorithm, ensuring a balanced distribution. In embracing this strategy, SwAV learns invariant features that encapsulate crucial semantic information.

Information maximization-based SSL approaches strive to learn representations by maximizing the invariance of corresponding features while minimizing the covariance between different features. Noteworthy methods in this realm include VICReg \cite{bardes2021vicreg} and Barlow Twins \cite{zbontar2021barlow}. VICReg specifically enforces feature representation invariance by amplifying their variance and curtailing the covariance between different features. This strategy ensures that the learned representations are not only informative but also capture the core attributes of the data. Contrasting with VICReg, which explicitly maximizes the variance of individual features, Barlow Twins centers its focus on minimizing cross-correlations across feature dimensions. It achieves this by mitigating the cross-correlation between the outputs of twin networks, each processing a distinct augmentation of the same image while concurrently accentuating the invariance of matching features.

\subsection{SSL in Pathology}

The advent of whole-slide scanners has enabled the digitization of histopathological slides, gradually transforming the field of anatomical pathology into a data-abundant domain. Recognizing this, SSL techniques are being increasingly employed in computational pathology to take advantage of the abundance of unlabeled data. In conjunction with smaller labeled datasets, these techniques promise to elevate diagnostic precision and bolster predictive modeling.

In the realm of computational pathology, SSL is gradually gaining adoption as a means to tackle challenges such as acquiring annotations for pathology slides, managing high-resolution images, and addressing the substantial variability in their features. Numerous studies have leveraged SSL for extracting features from WSIs and have utilized these features to achieve promising results in downstream histopathology image analysis tasks~\cite{chen2022scaling,dipalma2023histoperm,jiang2023masked,li_dual-stream_2021}. With the escalating adoption of SSL in computational pathology, there is a growing necessity to determine the applicability of general self-supervised methods to histopathology images. A recent benchmarking study~\cite{kang2023benchmarking} gauged various SSL methods across diverse pathology datasets for various downstream tasks, such as classification and nuclei instance segmentation tasks. Their results indicate that SSL can considerably uplift the performance on downstream tasks on histopathology images compared to ImageNet pre-trained and supervised models, especially when labeled data is scarce.

%% file: method.tex
\section{Method}

We recap SimSiam and then present our method for self-supervised representation learning and clustering.

\subsection{Preliminaries: SimSiam}
Self-supervised visual representation learning is a method for learning an embedding function that maps an input image $x$ to a representation. This is typically achieved by using a similarity measure designed to enforce similarity between augmented views. Starting with sets of data transformations, $\mathcal{T}_1$ and $\mathcal{T}_2$, we randomly sample transformations $t_1, t_2 \sim \mathcal{T}_1, \mathcal{T}_2$ and produce augmented views $x_1 = t_1 (x)$ and $x_2 = t_2(x)$. An encoder $f$ is used to produce representations $y_1 = f(x_1)$ and $y_2 = f(x_2)$, which are then fed to a projector $h$ to produce projections $z_1 = h(y_1)$ and $z_2 = h(y_2)$. As in SimSiam, we pass $z_1$ through a predictor $g$ to produce the prediction $p_1 = g(z_1)$. Additionally, we swap the views and produce a symmetric loss as follows:
\begin{equation}
    \mathcal{L}_{inv} = \text{sim}(p_1, \text{sg}(z_2)) + \text{sim} (p_2, \text{sg}(z_1))
\end{equation}
where
\begin{equation}
    \text{sim}(p, z) = - \frac{p}{\lVert p \rVert_2} \cdot \frac{z}{\lVert z \rVert_2},
\end{equation}

\noindent $\lVert \cdot \rVert$ is the $\ell_2-$norm, and $\text{sg}(\cdot)$ is the stop-gradient operation to prevent collapse.

\subsection{CluSiam: Cluster-Constrained SSL}
We build upon the SimSiam architecture by adding a cluster assigner $q$ that operates on the projections produced by $h$. We use the outputs of $h$ as the inputs to $q$ because the batch normalization layers ($\mathcal{BN}$) in $h$ stabilize the distribution of $q$'s inputs. We also introduce $\mathcal{BN}$ in $q$ because controlling the input scale is crucial for generating cluster assignment probabilities using $\softmax$. Given batches of views $X_1, X_2$, we produce projections $Z_1 = \{ h(f(x_i^{(1)})) : x_i^{(1)} \in X_1 \}, Z_2 = \{ h(f(x_i^{(2)})) : x_i^{(2)} \in X_2 \}$. We then concatenate these projections to obtain $Z = \text{concat}(Z_1, Z_2)$. The cluster assigner $q$ maps the concatenated projections $Z$ to cluster representations $A$ defined as:
\begin{equation}
    A = \frac{e^{(a_i) / \tau}}{\sum_i e^{(a_i) / \tau}} \in \mathbb{R}^{2 \cdot N \times K} \label{NormAssigns}
\end{equation}

\noindent where $A_{ij}$ is the element at the $i$-th row and $j$-th column of $A$, $\tau$ is the temperature, and $K$ is the exploration space that represents the maximum number of clusters allowed during the training. This operation is applied along each row, meaning that for every data point $i$, the sum of $A_{ij}$ over all clusters, $j$'s, equals 1. Finally, we map the cluster representation to the clusters as follows:
\begin{equation}
    C=\nonzero\left(\frac{\argmax(A^T)\cdot \sg(Z)}{\lVert \argmax(A^T) \cdot \sg(Z)\rVert_2}\right) \in \mathbb{R}^{k \times D} \label{CalCenter}.
\end{equation}
Notably, $\argmax(A)$ is not differentiable, so there is no real gradient for this operation. We approximate the gradient similar to the straight-through estimator and just copy gradients from $A$ to $\argmax(A)$, $\nabla_{\argmax(A)} C = \nabla_{A} C$, making it possible for backpropagation. $\sg(\cdot)$ denotes the stop-gradient operation, $\nonzero(\cdot)$ filters out vectors along the row dimension that are all zeros, and $k$ represents the count of non-zero centroids. The dimension $D$ corresponds to the feature dimension and is consistent with the dimensions of $p$ and $z$.

In our clustering module, unlike common clustering methods that use inter-class similarity or other SSL techniques that focus on the invariance between two different views, we do not impose any restrictions on inter-class similarity or the assignments between two different views, $a_{i}$ and $a_{n+i}$. Our cluster loss is solely defined by inter-cluster separation. This separation is defined as: 
\begin{equation}
\mathcal{L}_{cluster}=-\frac{\sum_{i}\sum_{i\neq j}(C^T C)_{ij}}{2\mathcal{C}^{2}_{k}}. \label{ClusterLoss}
\end{equation} 
Importantly, all the tensors in (\ref{ClusterLoss}) are $\ell_2$ normalized, so equation (\ref{ClusterLoss}) can be interpreted as representing the average cosine similarities between clusters. Furthermore, the stop-gradient operation is applied to all elements in (\ref{CalCenter}) to prevent the cluster assigner from collapsing to a trivial solution where all samples are assigned to the same set or cluster. We can view $A={a_1,\cdots,a_{2n}}$ as a latent variable, with our goal being to minimize $\mathcal{L}(A,Z)$. This optimization problemcan be solved by an alternating algorithm that fixes one set of variables and solves for the other set.

The intra-class similarity term was not introduced in this design because we neither use a contrastive formulation loss function like SimCLR nor an additional projection head to introduce the knowledge distillation architecture like BYOL. Directly optimizing for high intra-class similarity is prone to collapsing all samples into the same representation, which is a trivial solution and does not capture any meaningful information.

However, our cluster learning is still prone to collapse due to the presence of the $\softmax$ and $\argmax$ functions. The hard cluster assignment of the $\softmax$ and $\argmax$ functions limits the model to updating only the maximum scoring cluster during training. This does not encourage exploration of different combinations of cluster assignments, and other potential cluster assignments are left out of the backpropagation updates. 
As a result, the model is prone to getting stuck in a trivial single-cluster solution during training, a phenomenon known as ``collapsing''. This is especially likely as the number of valid clusters decreases. Using $\softmax$ and $\argmax$ based cluster assignment can exacerbate this problem, leading to the rapid collapse of a single cluster. This prevents the model from learning the intended cluster structures and, in turn, leads to meaningless representations since the cluster loss cannot be effectively optimized with only a single active cluster.

The task of cluster assignment can be viewed as a decision-making process in which the cluster assigner determines which samples belong in the same cluster. Striking a balance between exploration and exploitation is crucial in this decision-making process. Exploration refers to trialing different actions to learn more about the environment and their associated losses, while exploitation refers to choosing the action currently known to possess the lowest expected loss. To prevent the collapse of clustering due to continual updates only to the neuron with the highest probability value, we introduce randomness into the decision-making process by adding Gumbel noise ~\cite{jang2016categorical}. Gumbel noise is a random variable sampled from a Gumbel distribution. It proves useful in discrete action spaces, where a model must choose between a finite set of actions, as in our cluster assignment task. By replacing (\ref{NormAssigns}) with (\ref{GumbelNormA}), the cluster assigner can explore different cluster combinations based on their probabilities, thereby learning more about different cluster combinations and their associated losses. 
\begin{equation}
A = \frac{e^{(a_i+g_i)/\mathcal{\tau}}}{\sum_{i=1} e^{(a_i+g_i)/\mathcal{\tau}}}, \ g_i\sim G(0,1) ,\label{GumbelNormA} 
\end{equation} where $G$ represents the Gumbel distribution. 

Therefore, CluSiam can be trained effectively using the composite loss function, as shown in equation \eqref{FinalLoss}. In this equation, $\alpha$ serves as a hyperparameter, adjusting the magnitude of the weights. In our implementation, we set $\alpha=0.5$ by default. 
\begin{equation}
    \mathcal{L}_{CluSiam} = (1-\alpha) \cdot \mathcal{L}_{inv} + \alpha \cdot \mathcal{L}_{cluster}
    \label{FinalLoss}
\end{equation}

%% file: experiments_results.tex
\section{Experiments}

\begin{figure*}[!ht]
    \centering
    \includegraphics[width=\linewidth]{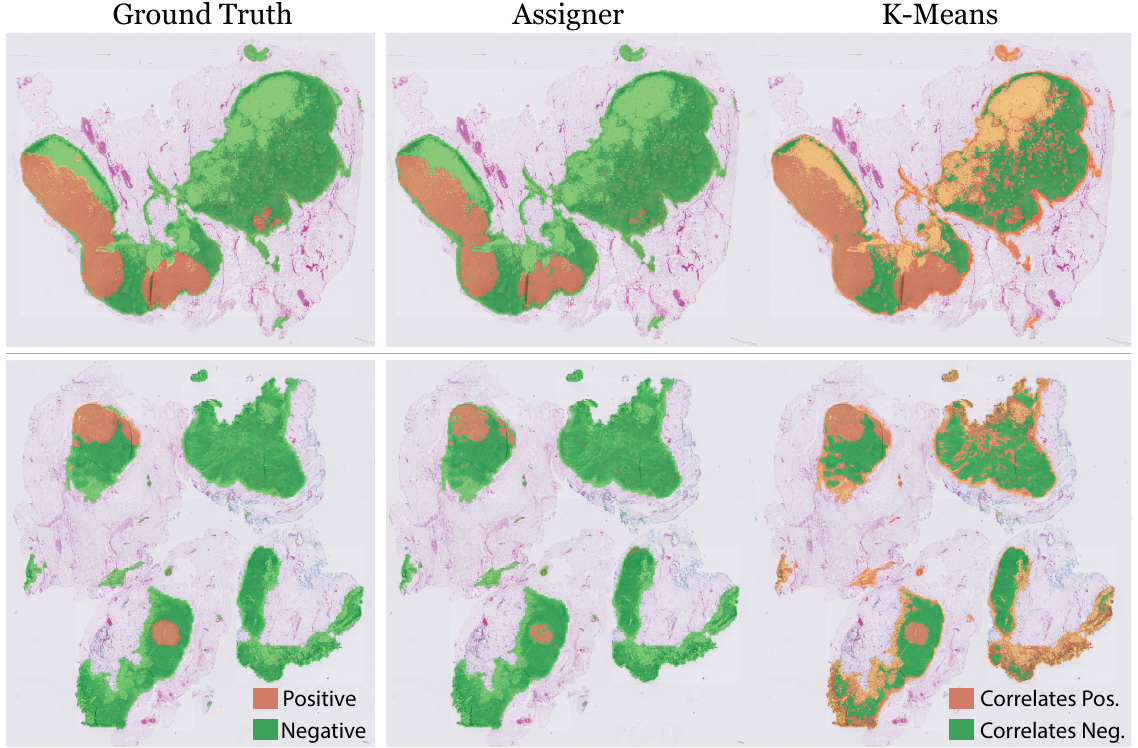}
    \caption{Cluster visualization for the unseen test set. Uncolored regions were filtered out during the preprocessing stage.}
    \label{fig:cluster}
\end{figure*}

In our experiments, we evaluated the performance of our proposed model on two clinically relevant whole-slide image datasets: the Camelyon16 \cite{litjens20181399} and the Pancreatic Cancer dataset~\cite{wu2023graph}. To extract representative image patches from the WSIs, we first removed the background and employed a sliding window technique to generate patches of size 224$\times$224 at a 20$\times$ magnification level (\SI{0.5}{\micro\meter}/pixel) from the tissue regions of a slide, with no overlap between patches. The Camelyon16 dataset is a publicly available dataset designed for the task of metastasis detection in breast cancer. It includes two classes, positive and negative slides, and consists of 270 training images and 129 testing images. After applying our patch extraction algorithm, we obtained approximately 2.6 million training and 1.2 million testing patches at 20$\times$ magnification for this dataset. The Pancreatic Cancer dataset includes three classes: negative (background class), neoplastic, and positive. This dataset includes 104 training slides and 39 testing slides, yielding approximately 300,000 training and 83,000 testing patches.

In our study, we compared our proposed CluSiam method against a supervised model and several commonly used SSL architectures as baselines. All SSL models, as well as the supervised model, were trained using a ResNet18~\cite{he2016deep} backbone for 50 epochs with a batch size of 512. The hyperparameters for training were set to be as identical as possible to their default values specified in the original studies with comparable settings. For the supervised model, we trained a patch classifier using cross-entropy loss. Due to the extreme class imbalance between positive and negative patches, we simply reweighted the loss function to avoid bias towards predicting the negative class in the supervised training. All features were extracted as 512-dimensional vectors from the ResNet18 backbone. The detailed hyperparameter settings can be found in the appendix. It's important to note that different architectures often incorporate varying image augmentation techniques, optimizers, and hyperparameters. As a result, comparisons between baseline methods may not be entirely fair due to inherent configuration differences. Our approach (CluSiam) used identical hyperparameter settings as SimSiam, which allows for a direct and equitable comparison between these two architectures. As a metric for evaluation, we employed two downstream tasks: clustering and classification.

\subsection{Clustering}
In the clustering task, we evaluated the performance of clustering algorithms using the Rand Index (RI). The performance was compared using different representations and clustering algorithms (Table~\ref{tab:cluster}). Importantly, our cluster assigner's output is different from traditional methods like K-Means. Unlike K-Means, which provides a hard cluster assignment for each patch, our assigner outputs a K-dimensional vector. This structure allows for more flexibility in generating cluster assignments beyond the typical use of $\argmax$. Such probabilistic outputs grant our method greater adaptability in clustering, especially when contrasted with the rigid assignments derived from K-Means. We visualized the cluster assignments of two WSIs generated simply using $\argmax$, alongside their respective ground truths, in Figure \ref{fig:cluster}.

\begin{table}[htbp]
  \centering
    \begin{tabular}{cc|cc}
    \hline
    Encoder & Cluster & Camelyon16 & Pancreatic \\
    \hline
    SimSiam & K-Means & 0.509 & 0.329 \\
    CluSiam & K-Means & 0.538 & 0.357 \\
    CluSiam & Assigner & \textbf{0.897} & \textbf{0.569} \\
    \hline
    \end{tabular}%
\caption{Rand Index on the testing set.}
\label{tab:cluster}
\end{table}%

\subsection{Classification}
For the classification task, we evaluated model performance using accuracy and Area Under the ROC Curve (AUC) metrics. We aggregated patch-level predictions to slide-level predictions using two multiple-instance learning techniques: Max-Pooling (Max) and Dual-Stream Multiple-Instance Learning (DSMIL). Given that optimal hyperparameters for the MIL models may differ based on the representations learned by various SSL methods, we conducted grid searches over learning rates [1e-2, 1e-3, 1e-4] and weight decays [1e-2, 1e-3, 1e-4], yielding 9 hyperparameter combinations per MIL model. This ensured a fair evaluation after optimizing each method's settings. The MIL models were trained for 50 epochs with a 5-epoch warmup and cosine annealing learning rate schedule. To select the best checkpoint for each representation, we split the training sets into 75\% training and 25\% validation partitions. The checkpoints with the highest validation set performance were chosen for final evaluation on the held-out test set (Table~\ref{tab:classCam} and ~\ref{tab:classPan}).

\begin{table}[htbp]
  \centering
    \begin{tabular}{cc|cc}
    \hline
    \multirow{2}[2]{*}{Agg.} & \multirow{2}[2]{*}{Rep.} & \multirow{2}[2]{*}{Accuracy} & \multicolumn{1}{c}{AUC} \\
          &       &         & Positive \\
    \hline
    \multirow{8}[2]{*}{Max} & Supervised & 0.628 & 0.501 \\
          & SimCLR & 0.860 & 0.951 \\
          & SwAV  & 0.853  & 0.845 \\
          & PCL   & 0.496  & 0.510 \\
          & Barlow. & 0.868  & 0.941 \\
          & BYOL  & 0.659  & 0.834 \\
          & SimSiam & 0.690  & 0.680 \\
          & CluSiam & \textbf{0.884}  & \textbf{0.952} \\
    \hline
    \multirow{8}[2]{*}{DSMIL} & Supervised & 0.651  & 0.635 \\
          & SimCLR & 0.822  & 0.874 \\
          & SwAV  & 0.876  & 0.859 \\
          & PCL   & 0.488  & 0.496 \\
          & Barlow. & 0.860  & 0.945 \\
          & BYOL  & 0.558  & 0.586 \\
          & SimSiam & 0.721  & 0.680 \\
          & CluSiam & \textbf{0.907}  & \textbf{0.952} \\
    \hline
    \end{tabular}%
\caption{Results on Camelyon16 dataset. The magnification level is \SI{0.5}{\micro\meter}/pixel. All the representations were trained using a batch size of 512 and the ResNet18 architecture.}
\label{tab:classCam}%
\end{table}%

\begin{table}[htbp]
  \centering
    \begin{tabular}{cc|cccc}
    \hline
    \multirow{2}[2]{*}{Agg.} & \multirow{2}[2]{*}{Rep.} & \multirow{2}[2]{*}{Acc.} & \multicolumn{3}{c}{AUC} \\
          &       &       & Neg.  & Neo.  & Pos. \\
    \hline
    \multirow{8}[2]{*}{Max} & Supervised & 0.359 & 0.313 & 0.562 & 0.494 \\
          & SimCLR & 0.462 & 0.565 & 0.549 & 0.720 \\
          & SwAV  & 0.462 & 0.497 & 0.451 & 0.726 \\
          & PCL   & \textbf{0.692} & 0.556 & 0.935 & 0.731 \\
          & Barlow. & 0.538 & 0.438 & 0.509 & 0.843 \\
          & BYOL  & 0.462 & 0.314 & 0.719 & 0.589 \\
          & SimSiam & 0.359 & 0.598 & 0.531 & 0.694 \\
          & CluSiam & 0.641 & \textbf{0.598} & \textbf{0.966} & \textbf{0.851} \\
    \hline
    \multirow{8}[2]{*}{DSMIL} & Supervised & 0.356 & 0.296 & 0.617 & 0.529 \\
          & SimCLR & 0.718 & 0.497 & 0.904 & 0.700 \\
          & SwAV  & 0.538 & 0.669 & 0.840 & 0.726 \\
          & PCL   & 0.744 & 0.710 & 0.969 & 0.797 \\
          & Barlow. & 0.615 & 0.672 & 0.957 & 0.831 \\
          & BYOL  & 0.564 & 0.527 & 0.981 & 0.697 \\
          & SimSiam & 0.590 & 0.686 & 0.910 & 0.803 \\
          & CluSiam & \textbf{0.769} & \textbf{0.754} & \textbf{0.985} & \textbf{0.883} \\
    \hline
    \end{tabular}%
\caption{Results on the Pancreatic Cancer dataset. The magnification level is \SI{0.5}{\micro\meter}/pixel. All the representations were trained using a batch size of 512 and the ResNet18 architecture.}
\label{tab:classPan}%
\end{table}%

\subsection{Ablation Study}

To investigate the influence of the core components of our CluSiam model, we conducted an ablation study on the Camelyon16 dataset. We created and trained several model variants with different combinations of the invariance loss term, cluster loss term, and Gumbel noise to evaluate their impacts.

\begin{table}[htbp]
  \centering
  \resizebox{0.99\linewidth}{!}{%
  \begin{tabular}{c|ccccc}
    \hline
    \multirow{2}{*}{Model} & \multirow{2}{*}{$\mathcal{L}_{inv}$} & \multirow{2}{*}{$\mathcal{L}_{cluster}$} & \multirow{2}{*}{Noise} & \multirow{2}{*}{Acc.} & \multicolumn{1}{c}{AUC} \\
    & & & & & Pos. \\ 
    \hline
    SimSiam & \checkmark     & -     & -     &0.721   &0.680  \\
    Cluster & -     & \checkmark     & \checkmark     &0.426  &0.627  \\
    Joint & \checkmark     & \checkmark     & -     &0.667  &0.563  \\
    CluSiam & \checkmark     & \checkmark     & \checkmark     &0.907  &0.952  \\
    \hline
    \end{tabular}
    }
\caption{Ablation study of CluSiam components on Camelyon16 using DSMIL aggregation.}
\label{tab:ablation}
\end{table}%

The stop gradient operation is crucial to prevent our model from collapsing, similar to its role in SimSiam. Additionally, incorporating Gumbel noise when assigning clusters is critical. Without this noise, the cluster assigner will collapse early in training, assigning all samples to one cluster. This collapsed state becomes equivalent to SimSiam training, as a single cluster removes the off-diagonal elements required to calculate the cluster loss $\mathcal{L}_\text{cluster}$. The gradient from $\mathcal{L}_\text{cluster}$ thus becomes zero in this situation. However, by introducing randomness in cluster assignment, the noise prevents updates from concentrating solely on the most probable cluster dimensions, thereby preventing early collapse. Together, the stop gradient and noise allow our model to escape these trivial single-cluster solutions, enabling effective joint optimization of $\mathcal{L}_\text{inv}$ and $\mathcal{L}_\text{cluster}$. As evidenced in Table~\ref{tab:ablation}, CluSiam outperforms the other models. We observed a collapse in the joint model's cluster assigner, with one dimension consistently dominating the others. This is likely because only the largest dimension receives updates, continually amplifying its magnitude and leading to a single dominant cluster. The joint model was expected to mirror SimSiam's performance, as its training dynamics should be identical to SimSiam after the assigner's collapse. However, its performance was inferior compared to SimSiam. This discrepancy may originate from the cluster assigner offering suboptimal initialization for the subsequent pure SimSiam training before its collapse. 

In the aftermath of our ablation studies on the new modules introduced to SimSiam, we delved into investigating the pivotal roles of original SimSiam components in ensuring training stability. A prime area of focus was the scale of inputs to the cluster assigner, as it might significantly influence this stability. The SimSiam projector, which interleaves $\mathcal{BN}$ layers between its linear layers and concludes with a $\mathcal{BN}$ layer, could be foundational for the clustering module's effectiveness. To empirically assess the role of controlled input scaling, we devised an experiment on two different projectors. The first projector, originating from SimSiam, concludes with a $\mathcal{BN}$ layer. The second projector, from BYOL, ends with a linear layer. We began by replacing the SimSiam-style projector with the BYOL variant in our CluSiam model, leading to the creation of the CluBYOL model with the cluster assigner module integrated into the BYOL architecture. The CluBYOL was initially trained using the BYOL-style projector. Subsequently, we utilized the SimSiam-style projector and undertook another training round for the CluBYOL model. Notably, both situations using the BYOL-style projector resulted in collapse, akin to the joint model depicted in Table~\ref{tab:ablation}, with a single cluster predominantly emerging. As highlighted in Table~\ref{tab:bn}, the SimSiam-style projector, characterized by its concluding with a $\mathcal{BN}$ layer, is instrumental in preventing such collapses.

\begin{table}[htbp]
  \centering
  \resizebox{1\linewidth}{!}{%
  \begin{tabular}{c|ccc}
    \hline
    \multirow{2}{*}{Model} & Projector& \multirow{2}{*}{Acc.} & \multicolumn{1}{c}{AUC} \\
    &Ending with $\mathcal{BN}$ &  & Pos. \\ 
    \hline
    \multirow{2}{*}{CluSiam}& -     &0.643    &0.591  \\
     & \checkmark     &0.907    &0.952  \\
    \multirow{2}{*}{CluBYOL}& -     &0.627   &0.658  \\
     & \checkmark     &0.923   &0.975  \\
    \hline
    \end{tabular}
    }
\caption{Results on Camelyon16 using DSMIL aggregation.}
\label{tab:bn}
\end{table}%

\begin{figure}[ht]
    \centering
    \begin{subfigure}{0.45\textwidth}
        \includegraphics[width=\textwidth]{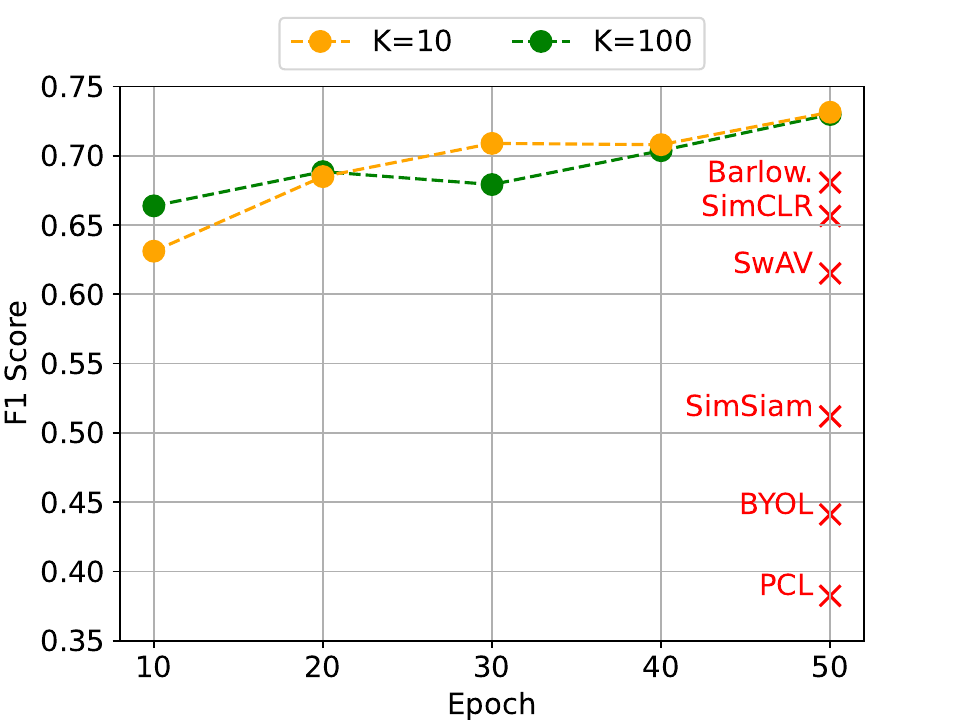}
        \caption{Patch-level F1 scores using the Top-1 KNN classifier.}
    \end{subfigure}
    \\
    \begin{subfigure}{0.45\textwidth}
        \includegraphics[width=\textwidth]{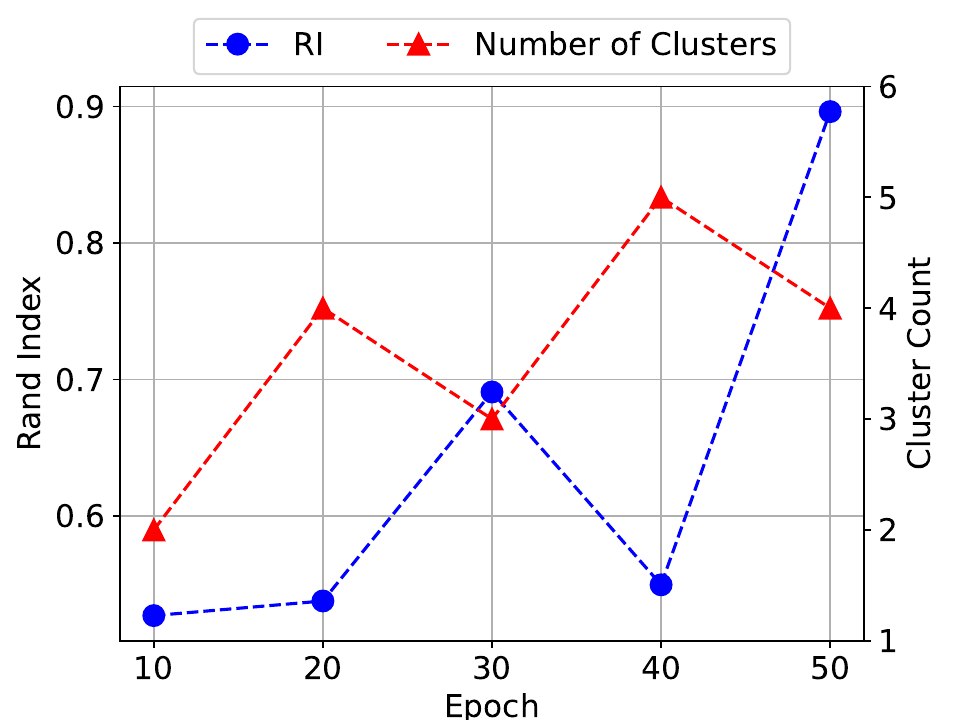}
        \caption{WSI-level clustering patterns with exploration space $K$=10.}
    \end{subfigure}
    \\
    \begin{subfigure}{0.45\textwidth}
        \includegraphics[width=\textwidth]{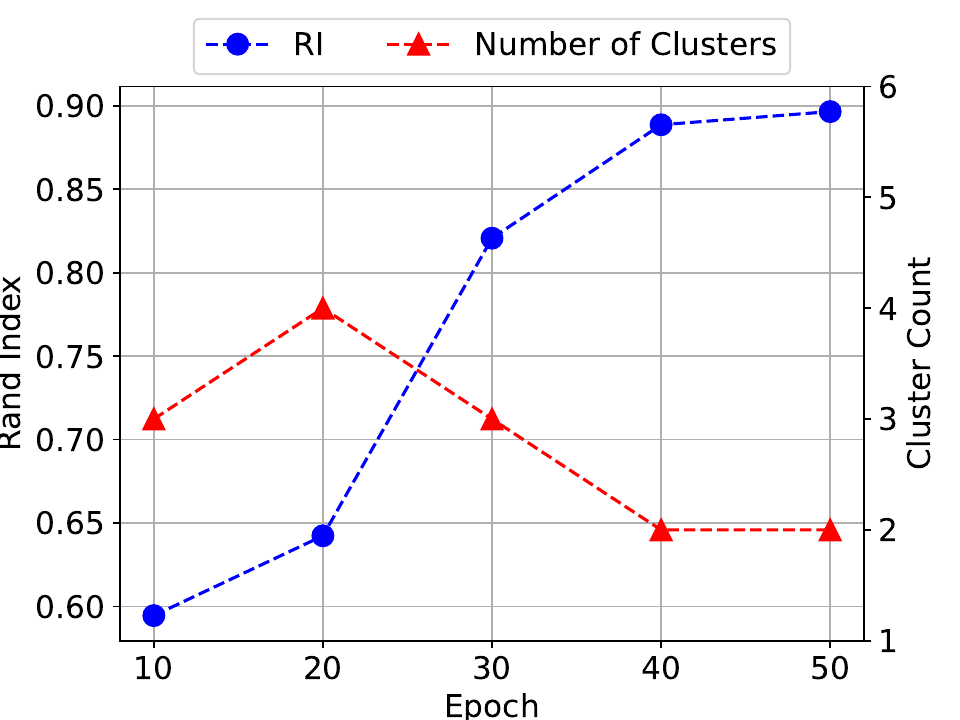}
        \caption{WSI-level clustering patterns with exploration space $K$=100.}
    \end{subfigure}
    \caption{Comparison of models with different exploration spaces $K$ on Camelyon16.}
    \label{fig:diff k}
\end{figure}

To further analyze the behavior of the clustering module, we examined the impact of the exploration space size, denoted as $K$. When Gumbel noise is introduced, some samples exhibit the highest probability of remaining in the cluster with the highest output value, while also possessing a high probability of transitioning to a nearby or similar cluster. Some samples that are difficult to differentiate might be allocated with near-equal probabilities across multiple centroids. A larger value of $K$ results in more refined clustering. Conversely, when $K$ is small, achieving definitive clustering becomes difficult. For instance, with $K$ set to 3, the cluster assigner can allocate some hard-to-distinguish samples into a third cluster, roughly equidistant from the first two. Yet, with $K$ set to 2, the assigner can only place samples into one of the two clusters. Importantly, when $K$ is 1, the model essentially becomes SimSiam since its loss function and backpropagation are equivalent to those in SimSiam. In this scenario, the cluster assigner lacks the flexibility to differentiate between samples by assigning them to different clusters.

In our experiments, we assessed the influence of $K$ on model behavior by training two models with exploration spaces of $K=10$ and $K=100$. Using a Top-1 KNN classifier's F1 score for patch-level performance evaluation, both models displayed comparable classification performance, as depicted in Figure \ref{fig:diff k}. Despite this similarity in classification, their clustering behaviors were distinct. The model with $K=10$ exhibited a larger number of clusters and greater fluctuations in both cluster counts and the Rand Index. In contrast, the larger exploration space of $K=100$ allowed the assigner to stabilize on more definitive assignments faster. This difference underlines the influence of $K$ in CluSiam's clustering. Specifically, both the number of clusters and the Rand Index fluctuate more with $K$=10. This limited exploration space prevents highly fine-grained cluster assignments. In contrast, a larger exploration space of $K$=100 provides more granularity for refined clustering actions, enabling the assigner to stabilize on more definitive assignments rapidly. The cluster count also becomes more consistent with $K$=100. These observations highlight the impact of $K$ on CluSiam's clustering dynamics.